\documentclass[11pt]{article}

\usepackage[preprint]{acl}

\usepackage{times}
\usepackage{latexsym}

\usepackage[T1]{fontenc}
\usepackage[utf8]{inputenc}

\usepackage{microtype}

\usepackage{inconsolata}

\usepackage{graphicx}
\usepackage{graphicx}
\usepackage{amsmath}
\usepackage{amssymb}
\usepackage{mathtools}
\usepackage{amsthm}

\usepackage{tcolorbox}
\tcbuselibrary{skins}
\usepackage{lipsum}
\usepackage{algorithm}
\usepackage{algorithmic}
\usepackage{marvosym}
\usepackage{booktabs}

\usepackage{amsmath,amssymb,amsthm}
\usepackage{listings}
\usepackage{enumitem}

\theoremstyle{plain}

\theoremstyle{definition}

\theoremstyle{remark}

\title{Adaptive Test-Time Compute Allocation via Learned Heuristics over Categorical Structure}

\author{Shuhui Qu \\
  Stanford University \\
  \texttt{shuhuiq@stanford.edu} \\
  }

\begin{document}
\maketitle
\begin{abstract}
Test-time computation has become a primary driver of progress in large language model (LLM) reasoning, but it is increasingly bottlenecked by expensive verification.
In many reasoning systems, a large fraction of verifier calls are spent on redundant or unpromising intermediate hypotheses.
We study reasoning under a \emph{verification-cost-limited} setting and ask how verification effort should be allocated across intermediate states.
We propose a state-level selective verification framework that combines (i) deterministic feasibility gating over a structured move interface, (ii) pre-verification ranking using a hybrid of learned state-distance and residual scoring, and (iii) adaptive allocation of verifier calls based on local uncertainty.
Unlike solution-level best-of-$N$ or uniform intermediate verification, our method distributes verification where it is most informative.
On the \textsc{MATH} benchmark, our approach achieves higher accuracy than best-of-$N$, majority voting, and beam search while using 44\% fewer verifier calls.
\end{abstract}

\section{Introduction}
\label{sec:intro}

Recent progress in LLM reasoning increasingly relies on \emph{test-time computation}: rather than emitting a single completion, systems sample multiple candidates, branch over intermediate steps, and apply explicit selection or verification to filter errors before committing to an answer \citep{wei2022chain,wang2022self,yao2022react,chen2024tree,teng2025atom}. This paradigm is particularly effective in multi-step symbolic domains where local mistakes compound, such as grade-school and competition mathematics \citep{cobbe2021training,hendrycks2021measuring,lewkowycz2022solving}.
At the same time, many strong reasoning pipelines now include an explicit \emph{verifier} to score or validate intermediate reasoning \citep{cobbe2021training,lightman2023let,zheng2023judging}.

However, scaling test-time compute introduces a deployment bottleneck: as the number of branches and checks increases, \emph{verification} often becomes the dominant marginal cost because it requires additional model calls and/or external tool execution across many intermediate decision points \citep{snell2024scaling,gao2023pal,chen2022program}. This motivates a cost-sensitive view of reasoning where the key scarce resource is not merely tokens, but the number of expensive \emph{verifier invocations} and executions one can afford per query \citep{snell2024scaling,lightman2023let}. In this setting, the central question is not whether extra compute helps (it does), but \emph{how to allocate it}.

The core problem is \emph{selective allocation} under branching: at an intermediate state, modern generators can propose many plausible next steps, but only a small fraction are both feasible and useful, especially on hard instances in \textsc{MATH} \citep{hendrycks2021measuring,lewkowycz2022solving}.
Common test-time scaling strategies allocate compute in coarse ways.
Solution-level approaches such as best-of-$N$ and self-consistency invest budget by generating many complete solutions and aggregating, without explicitly resolving ambiguous \emph{local} choices along a trajectory \citep{wang2022self}. Tree/graph search variants distribute compute over intermediate steps, but typically rely on fixed-width expansion or heuristic scoring, and may still waste budget on candidates that are ill-formed under the current constraints \citep{chen2024tree,borgeaud2022improving,teng2025atom}. In parallel, compute-optimal allocation work shows that \emph{problem-dependent} budgeting can substantially improve the accuracy--cost frontier \citep{snell2024scaling}, but it is largely posed at a \emph{problem-level} granularity and does not directly address heterogeneity \emph{within} a single problem, where easy and hard decision points can coexist in the same reasoning trace.

In this work we study \emph{verification-cost-limited reasoning} and ask:
\begin{quote}
\emph{Can we reduce expensive verifier calls without materially degrading correctness by allocating verification selectively at \textbf{intermediate states}?}
\end{quote}
We propose a \emph{gated competition} policy built on two principles.
First, we exploit \emph{deterministic feasibility structure} at the move interface: each candidate step is emitted as a structured operator with explicit arguments, enabling cheap gates that reject moves with explicit interface violations (e.g., parse/scope/type incompatibilities or contradictions with tracked constraints) \emph{before} any verifier calls.
Second, we allocate verification according to \emph{state-local uncertainty} rather than a uniform or purely problem-level rule.
Concretely, we (i) apply deterministic gating, (ii) rank surviving candidates with a hybrid pre-verification score that combines a learned distance-to-goal signal with a learned residual trained from verifier-labeled candidate lists (process-style supervision without replacing the verifier) \citep{cobbe2021training,lightman2023let,qu2025sqbcp}, and (iii) choose a \emph{state-conditional} verification budget $k(w)$ using a local uncertainty proxy so that ambiguous branching points receive more verification while clear decisions receive less \citep{snell2024scaling}.

We evaluate on \textsc{MATH}, a standard testbed that stresses long-horizon symbolic reasoning and is widely used to study test-time compute and search-based reasoning \citep{hendrycks2021measuring,lewkowycz2022solving}.
We compare against best-of-$N$, majority voting, and beam search under matched generation budgets, reporting verifier calls as the primary cost metric. Across budgets, our method improves the call-efficiency frontier, showing that \emph{how} verification is distributed across intermediate states can materially change the accuracy achieved under a fixed verifier-call budget.

\paragraph{Contributions.}
\begin{enumerate}[leftmargin=*,itemsep=2pt]
\item We formalize intermediate-step reasoning under a \emph{verifier-call} cost model and introduce a gated competition policy that combines deterministic feasibility filtering, hybrid pre-verification scoring, and state-conditional verification allocation \citep{snell2024scaling,cobbe2021training,lightman2023let}.
\item We provide a practical training recipe for a \emph{pre-verification} residual scorer from verifier-labeled candidate lists (within-state ranking) with an optional trajectory signal, explicitly targeting top-$k$ selection under a call budget \citep{lightman2023let,zheng2023judging}.
\item On \textsc{MATH}, we demonstrate improved accuracy--cost trade-offs relative to best-of-$N$, majority voting, and beam search.
\end{enumerate}

\section{Related Work}
\label{sec:related_work}

\paragraph{Test-time compute scaling for reasoning.}
A large body of work improves LLM reasoning by increasing test-time computation via sampling, search, and aggregation.
Chain-of-Thought prompting elicits intermediate rationales that improve multi-step accuracy \citep{wei2022chain}, while self-consistency aggregates multiple sampled rationales/answers to reduce variance \citep{wang2022self}.
Beyond pure sampling, several lines explicitly branch over intermediate states, including algorithmic tree search for language-model agents and reasoning-time search controllers \citep{chen2024tree,teng2025atom}.
Program-guided variants offload parts of reasoning to execution, reducing arithmetic and symbolic errors through tool-based computation \citep{gao2023pal,chen2022program}.
These methods establish that more test-time compute helps, but do not by themselves specify how to allocate \emph{verification} cost across intermediate branching.

\paragraph{Verification, process supervision, and LLM judges.}
Verifier-centric pipelines train or use models that score correctness of solutions or intermediate steps.
Early work trained verifiers for math word problems and showed that reranking/selection with a verifier can substantially boost accuracy \citep{cobbe2021training}.
Process supervision and step-level reward models provide dense feedback over reasoning traces and can guide selection among candidates \citep{lightman2023let}.
Separately, LLM-as-a-judge has emerged as a practical evaluation and selection primitive for open-ended generations \citep{zheng2023judging}.
Our work is aligned with verifier-based selection, but focuses on \emph{where} to spend verifier calls: we treat learned scoring as \emph{pre-verification} prioritization and rely on explicit verification for acceptance.

\paragraph{Adaptive compute and budget allocation.}
Rather than spending a fixed budget everywhere, recent work studies how to allocate test-time compute to maximize accuracy per unit cost.
Compute-optimal allocation demonstrates strong gains from \emph{problem-dependent} budgeting and analyzes several allocation primitives, including lookahead-style intermediate exploration \citep{snell2024scaling}.
% We share the objective of improving the accuracy--cost frontier, but differ in \emph{granularity and cost model}: we treat \emph{verifier calls} as the dominant marginal cost and allocate verification at \emph{intermediate states}, targeting within-problem heterogeneity that problem-level budgeting does not directly resolve.

\paragraph{Structured generation, feasibility filtering, and constrained decoding.}
A complementary direction is to reduce wasted computation by enforcing structure during generation.
Constrained decoding and grammar-based generation restrict outputs to syntactically valid forms, improving reliability in structured tasks \citep{hokamp2017lexically,post2018fast}.
Tool-use frameworks also emphasize structured interfaces (actions with arguments) to make downstream execution and validation well-defined \citep{schick2023toolformer}.
Similarly, we enforce a structured move interface and use cheap, deterministic feasibility checks to eliminate clearly invalid candidates \emph{before} invoking an expensive verifier.

\paragraph{Retrieval and reuse in reasoning.}
Retrieval-augmented generation improves factuality and reasoning by conditioning generation on retrieved context \citep{lewis2020retrieval,borgeaud2022improving}.
Large-scale retrieval inside language modeling (e.g., RETRO-style approaches) further demonstrates that retrieval can substitute for some parametric memory and improve generation quality \citep{borgeaud2022improving}.
In contrast, our retrieval component (when enabled) targets \emph{operator/move reuse} rather than external knowledge retrieval; it is most closely related to reusing verified sub-steps within a structured search space.

\paragraph{Contextual bandits and resource allocation.}
Our setting is also conceptually adjacent to contextual bandits and sequential resource allocation: at each state, the system must choose which candidates to evaluate under a limited budget, using context-dependent uncertainty to decide where evaluation has high marginal value\citep{li2010contextual}.
However, we do not cast training or inference as a bandit algorithm, since we have supervised verifier labels from exploration logs and our primary objective is improved verification-call efficiency under a fixed generator--verifier interface.
We therefore treat bandit-style allocation as related conceptual framing rather than a direct baseline in our experiments.

\section{Problem Formulation: verification-cost-limited reasoning}
\label{sec:problem_setting}

We study intermediate-step reasoning under a \emph{verification-call cost model}.
Given an input problem $x$, a reasoning system repeatedly proposes candidate next-step updates and may invoke a verifier to check step validity. We treat verifier invocations as the dominant marginal cost, and aim to \emph{minimize verifier calls subject to high task accuracy}.

\paragraph{State.}
We represent a reasoning \emph{state} as
$
w=(r,s,\ell),
$
where $s$ is the current symbolic trace (e.g., equations or partial derivation), $\ell$ is an explicit constraint context (e.g., known bindings, domain constraints, tracked invariants), and $r$ is the remaining verification budget.
We omit an explicit time component; time is subsumed into $r$ as a budgeted resource.

\paragraph{Moves and structured interface.}
At state $w$, the system forms a candidate set of \emph{moves}
\begin{equation}
\mathcal{M}(w) \;=\; \mathcal{M}_{\text{gen}}(w)\ \cup\ \mathcal{M}_{\text{ret}}(w),
\label{eq:move_union}
\end{equation}
where $\mathcal{M}_{\text{gen}}(w)=\mathrm{Gen}(w)$ are newly proposed moves and $\mathcal{M}_{\text{ret}}(w)$ are retrieved moves (e.g., cached operators) whose preconditions match the current context.
Each move is a structured hypothesis
$
m=(\texttt{op},\texttt{args})
$
that can be deterministically parsed into an operator with well-defined arguments.
A move induces an updated state via a deterministic application operator
$
w'=\mathrm{Apply}(w,m).
$

This structured interface is essential: it provides a domain where deterministic feasibility checks are meaningful, without attempting to soundly analyze arbitrary free-form text.

\paragraph{Verifier.}
A verifier $\mathcal{V}(w,m)\in\{0,1\}$ returns whether applying $m$ is an acceptable reasoning step at $w$.
The verifier may be an LLM judge, tool execution, or any other expensive oracle; our cost metric counts verifier invocations.
We emphasize that \emph{gating} and \emph{verification} play different roles: gates cheaply rule out moves that violate explicit deterministic constraints, while the verifier adjudicates semantic correctness among remaining candidates.

\paragraph{Goal and objective.}
Let $\mathrm{Solve}(x)$ denote producing a final answer for problem $x$.
We seek policies that optimize the accuracy--cost frontier:
\begin{equation}
\begin{split}
    \max_{\pi}\ &\mathbb{E}\big[\mathbf{1}\{\mathrm{Solve}(x)\ \text{is correct}\}\big] \\
 \text{s.t.} &\quad \mathbb{E}\big[\#\text{verifier calls}\big] \le B.
\end{split}
\label{eq:budgeted_objective}
\end{equation}
We focus on policies that allocate verification \emph{at intermediate states}, where branching is large and uniform verification is expensive.

\section{Method}
\label{sec:method}

Our core hypothesis is that \emph{state-level verification allocation} yields better accuracy--cost tradeoffs than uniform verification, because local uncertainty varies widely across states within a single problem\footnote{Note, we do \emph{not} perform multi-step rollout evaluation to optimize a learned reward; our policy is single-step at each visited state and allocates \emph{verifier calls}, not rollout depth. This distinction matters empirically because lookahead-style PRM search can underperform budget-matched best-of-$N$ in prior work; our method targets a different bottleneck: \emph{which intermediate candidates receive expensive verification}.}.
We implement this as a three-stage \emph{gated competition} pipeline:
\begin{enumerate}
\item \textbf{Deterministic feasibility gates} reject moves that violate explicit necessary conditions under the structured interface (no verifier calls).
\item \textbf{Competitive scoring} ranks surviving moves using a hybrid heuristic that decomposes structural progress and a learned pre-verification residual.
\item \textbf{State-conditional allocation} chooses $k(w)$ moves to verify, spending more verification budget when the local decision is uncertain.
\end{enumerate}

\begin{figure}[t]
    \centering
    \includegraphics[width=\linewidth]{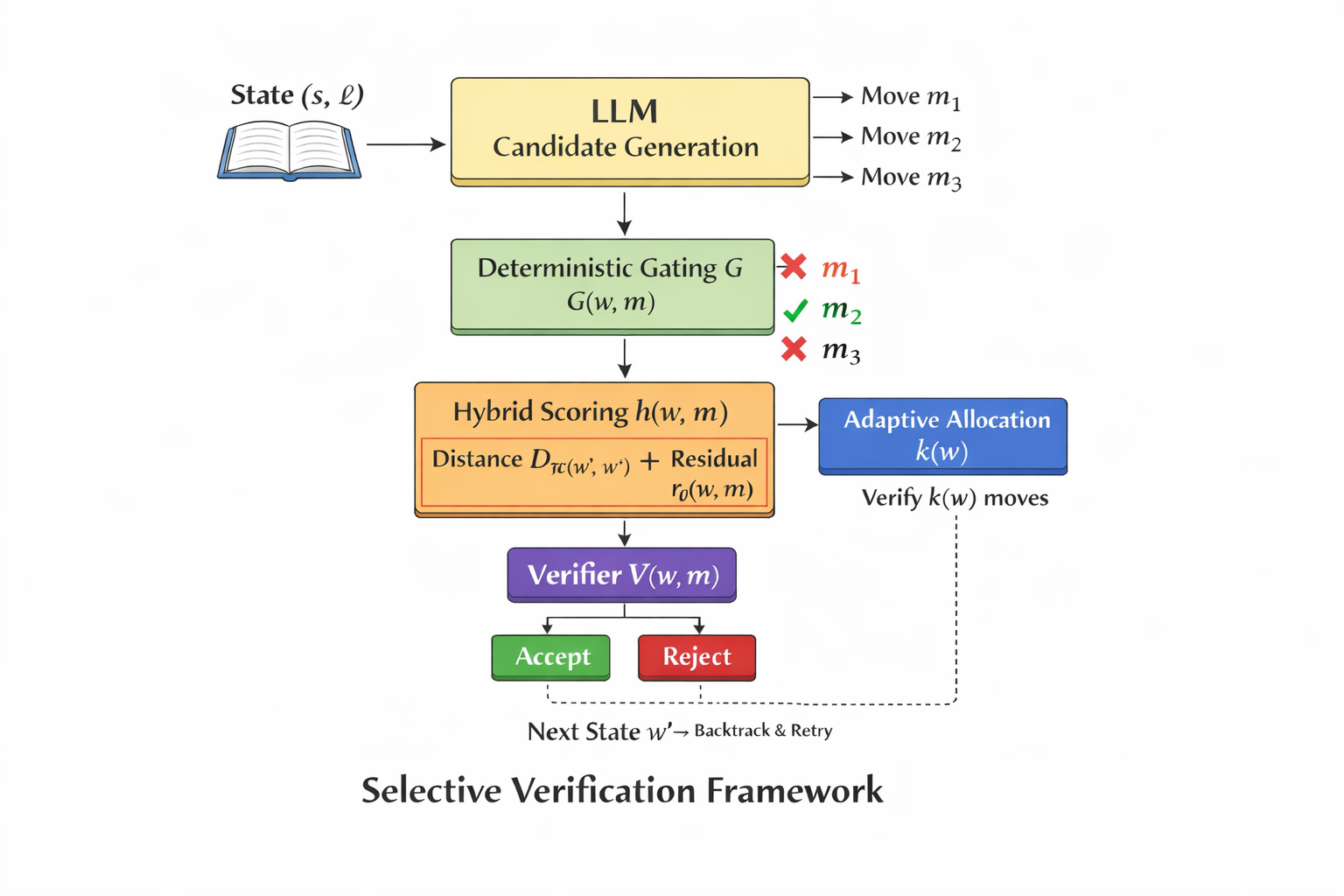}
    \caption{Overall framework.}
    \label{fig:framework}
\end{figure}

\subsection{Deterministic feasibility gates}
\label{sec:gates}

We apply two deterministic feasibility predicates and retain only moves that pass both:
\begin{equation}
\begin{split}
    \mathcal{M}^{\dagger}(w) \;=\; &\{m\in \mathcal{M}(w):  \\
    &\mathcal{G}_{\text{struct}}(w,m)=1 \ \wedge\ \mathcal{G}_{\text{ctx}}(w,m)=1\}
\end{split}
\label{eq:gated_moves}
\end{equation}

\textbf{Gate 1: structural compatibility.}
$\mathcal{G}_{\text{struct}}(w,m)$ rejects moves that fail deterministic checks such as parseability of $(\texttt{op},\texttt{args})$, symbol scope, and operator applicability to the current symbolic form in $s$.
These checks operate only on the structured move representation and the explicit context $\ell$.

\textbf{Gate 2: constraint compatibility.}
$\mathcal{G}_{\text{ctx}}(w,m)$ rejects moves that contradict the current constraint context $\ell$.
Let $\mathrm{Ctx}(w)$ denote extracted constraints from $w$ and let $\Delta(w,m)$ denote constraints newly introduced by applying $m$.
We define:
\begin{equation}
\begin{split}
    \mathcal{G}_{\text{ctx}}(w,m)=
\mathbf{1}\Big[
&\mathrm{Consistent}\big(\mathrm{Ctx}(w)\cup \Delta(w,m)\big)
\ \wedge\ \\
&\neg\mathrm{Violates}\big(\mathrm{Ctx}(w),\Delta(w,m)\big)
\Big].
\end{split}
\label{eq:ctx_gate}
\end{equation}

The gate does not guarantee global semantic soundness.
Instead, the gates are \emph{deterministic necessary-condition checks} with respect to the structured interface and the explicitly tracked constraints in $\ell$. Passing the gates does not imply correctness; failing the gates indicates an \emph{explicit violation} (e.g., undefined symbol, inapplicable operator type, direct contradiction with tracked invariants).

\subsection{Competitive scoring: structure + learned pre-verification residual}
\label{sec:scoring}

Among gated survivors, we rank moves using
\begin{equation}
\begin{split}
    h(w,m)\;&=\;D_{\text{type}}(w',w^\star)\;+\;r_\theta(w,m),\\
    w'&=\mathrm{Apply}(w,m),
\end{split}
\label{eq:hybrid_score}
\end{equation}
where $D_{\text{type}}$ is a structural distance between states and $r_\theta$ is a learned residual trained for \emph{top-$k$ selection under a verifier-call budget}.

\paragraph{Structural distance $D_{\text{type}}$}
We instantiate $D_{\text{type}}$ as a \emph{latent} structural distance computed from an LLM encoder.
Let $e(\cdot)$ denote the frozen (or lightly tuned) embedding function given by the generator backbone, applied to a canonical serialization of a state.
We compute embeddings for the post-move state and the goal specification:
\[
z_{w'} = e((w'),\qquad z_{w^\star} = e(w^\star).
\]
We then define
\begin{equation}
D_{\text{type}}(w',w^\star)
=
d(z_{w'},z_{w^\star}),
\label{eq:dtype_embed}
\end{equation}
where $d(\cdot,\cdot)$ is a distance metric in embedding space. While cosine distance is a simple choice, we find that a \emph{learned} distance function  performs better in our settings. Crucially, $D_{\text{type}}$ is \emph{not} used as a hard reachability gate; it serves only as a soft structural prior in Eq.~\eqref{eq:hybrid_score}, since approximate goal tests can be brittle.

\paragraph{Residual $r_\theta(w,m)$} predicts which \emph{feasible} moves are worth spending verifier calls on under a finite budget.
It does not accept or reject moves; acceptance is determined only by the verifier $\mathcal{V}(w,m)$.
Thus $r_\theta$ affects efficiency (and indirectly success under a fixed budget) by prioritizing verification, but cannot by itself introduce invalid steps.

\subsubsection{Training data from verifier-labeled candidate lists}
\label{sec:data}

We train $r_\theta$ from exploration runs on training problems using the same generate--gate--verify loop as at test time.
Each run visits a sequence of states; at each visited state $w$ we log:
(i) the post-gating candidate set $\mathcal{M}^{\dagger}(w)$, and
(ii) verifier labels for all candidates in that set:
\begin{equation}
y(w,m) \;=\; \mathcal{V}(w,m)\in\{0,1\}, \qquad m\in \mathcal{M}^{\dagger}(w).
\label{eq:verifier_label}
\end{equation}
This yields a dataset of \emph{state-wise candidate lists} with sparse positives, where positives are exactly the moves the verifier would accept under the same structured interface.
Optionally, when the system commits an accepted move, we also record the transition $(w,m,w')$ along the accepted trace for auxiliary analysis (e.g., downstream utility among multiple acceptable moves).

\subsubsection{Embedding model for $r_\theta$}
\label{sec:model}

We also parameterize $r_\theta$ as a lightweight scorer over LLM latent embeddings of the \emph{state} and the \emph{move}.
Let $e(\cdot)$ denote the same LLM embedding function used above, applied to serialized inputs.
We compute
\[
z_w = e(w),\qquad z_m = e(m),
\]
We then score with a small MLP head:
\begin{equation}
r_\theta(w,m)
=
g_\theta\!\big([z_w;z_m;z_{w^\star}]\big).
\label{eq:r_model_embed}
\end{equation}
Including $z_{w^\star}$ allows the residual to model instance-specific preferences relative to the goal, while $D_{\text{type}}$ provides a direct embedding-distance progress signal. In our instantiation, $e(\cdot)$ is frozen and only $g_\theta$ is trained.

\subsubsection{Loss: within-state ranking from verifier feedback}
\label{sec:loss}

Because $r_\theta$ is used for top-$k$ selection, we train it as a \emph{ranking} model within each state.
For each state $w$, define:
\begin{equation}
    \begin{split}
        &\mathcal{M}^+(w)=\{m\in \mathcal{M}^{\dagger}(w): y(w,m)=1\} \\
        &\mathcal{M}^-(w)=\{m\in \mathcal{M}^{\dagger}(w): y(w,m)=0\}.
    \end{split}
\end{equation}

We sample pairs $(m^+,m^-)$ with $m^+\in\mathcal{M}^+(w)$ and $m^-\in\mathcal{M}^-(w)$ and minimize the logistic pairwise ranking loss
\begin{equation}
\begin{split}
    \mathcal{L}_{\text{rank}}(\theta)
=&
\mathbb{E}_{w}\;
\mathbb{E}_{m^+\sim \mathcal{M}^+(w),\, m^-\sim \mathcal{M}^-(w)}\\
&\Big[
\log\big(1+\exp(r_\theta(w,m^+)-r_\theta(w,m^-))\big)
\Big].
\end{split}
\label{eq:rank_loss}
\end{equation}
Minimizing \eqref{eq:rank_loss} encourages verifier-acceptable moves to rank ahead of rejected moves.
(Equivalently, one can swap the sign convention; we adopt the convention that \emph{lower} $h(w,m)$ ranks higher.)

\paragraph{Optional trajectory signal (remaining-steps).}
Verifier labels provide strong \emph{local} supervision (which candidates are acceptable at a given state), but they may not distinguish among multiple acceptable moves by \emph{downstream} utility.
When we have accepted trajectories $\tau=(w_0,m_0,w_1,\dots,w_L)$ that reach a correct solution, we add a weak cost-to-go signal based on \emph{remaining steps}.
Let
\[
g(w_i)\;=\;L-i
\]
be the remaining number of accepted moves to termination along $\tau$.
We use $g(\cdot)$ as an extra supervision target that prefers moves that advance toward termination in fewer verified steps.
Concretely, we regress the residual toward this signal:
\begin{equation}
\mathcal{L}_{\text{traj}}(\theta)
=
\mathbb{E}_{(w_i,m_i)\in \tau}
\Big[
\big(r_\theta(w_i,m_i) - \alpha\, g(w_i)\big)^2
\Big],
\label{eq:traj_loss}
\end{equation}
where $\alpha$ rescales the step-count target to match the residual magnitude.
Our final objective is
\begin{equation}
\mathcal{L}(\theta)=\mathcal{L}_{\text{rank}}(\theta)+\lambda\,\mathcal{L}_{\text{traj}}(\theta),
\label{eq:training_loss}
\end{equation}
where $\mathcal{L}_{\text{rank}}$ is the primary signal and  $\mathcal{L}_{\text{traj}}$ is a weak signal.

\subsection{State-conditional verification allocation}
\label{sec:allocation}

Let $\mathcal{M}^{\dagger}(w)$ denote the move set after gating.
We compute scores $\{h(w,m): m\in \mathcal{M}^{\dagger}(w)\}$ and use their dispersion as a proxy for local decision uncertainty:
\begin{equation}
\sigma^2(w)=\mathrm{Var}\Big(\{h(w,m): m\in \mathcal{M}^{\dagger}(w)\}\Big).
\end{equation}
We then allocate a state-conditional verification budget
\begin{equation}
k(w)=\mathrm{clip}\Big(k_{\min},k_{\max}, k_{\text{base}}\big(1+\beta(\sigma(w)/\bar{\sigma}-1)\big)\Big),
\label{eq:kw}
\end{equation}
and verify only the top-$k(w)$ moves by $h(w,m)$.
When scores are concentrated, verifying many candidates is wasteful; when many candidates are close, additional verification yields higher marginal value.

\subsection{Algorithm}
\label{sec:algorithm}

\begin{algorithm}[t]
\caption{Gated competition with state-conditional verification}
\label{alg:main}
\small
\begin{algorithmic}[1]
\REQUIRE initial state $w_0=(B,s_0,\ell_0)$, goal specification $w^\star$
\FOR{$t=0,1,2,\dots$ until termination}
    \STATE Form candidate moves $\mathcal{M}(w_t)=\mathcal{M}_{\text{gen}}(w_t)\cup \mathcal{M}_{\text{ret}}(w_t)$
    \STATE $\mathcal{M}^{\dagger}(w_t)\leftarrow \{m\in\mathcal{M}(w_t): \mathcal{G}_{\text{struct}}(w_t,m)=1 \wedge \mathcal{G}_{\text{ctx}}(w_t,m)=1\}$
    \STATE Score each $m\in\mathcal{M}^{\dagger}(w_t)$ using $h(w_t,m)$ (Eq.~\eqref{eq:hybrid_score})
    \STATE Set $k_t\leftarrow k(w_t)$ (Eq.~\eqref{eq:kw}) and select top-$k_t$ moves
    \STATE Verify selected moves using $\mathcal{V}(w_t,m)$ \hfill ($k_t$ verifier calls)
    \STATE If any move is accepted, commit the best verified move and update $w_{t+1}\leftarrow \mathrm{Apply}(w_t,m)$
    \STATE Otherwise, retry at $w_t$ or backtrack)
\ENDFOR
\end{algorithmic}
\end{algorithm}

\paragraph{Connection to competition and global feedback.}
At each state, multiple candidates compete via $h(w,m)$ under a limited verification capacity $k(w)$.
Verifier outcomes provide high-precision feedback used to train $r_\theta$ via within-state ranking and to update any optional retrieval/cache mechanism, aligning with an intuition of parallel proposal, competitive selection, and global feedback.

\section{Experiments}
\label{sec:exp}

\subsection{Goal and empirical questions}
\label{sec:exp_goal}

We study \emph{verification-cost-limited} reasoning, where the dominant marginal test-time cost is invoking an expensive verifier $\mathcal{V}(w,m)$ on intermediate moves.
Prior work~\citep{snell2024scaling} demonstrates that test-time compute allocation at the \emph{problem level} can yield significant efficiency gains over uniform best-of-N by conditioning on problem difficulty.
Our scientific question is whether \emph{finer-grained, state-level} allocation provides additional benefits by combining (i) deterministic feasibility structure at the \emph{move} interface and (ii) uncertainty-aware \emph{state-level} allocation of verification effort.

\subsection{Setup}
\label{sec:setup}

\paragraph{Task and data.}
We evaluate on MATH~\citep{hendrycks2021measuring} using the 500-problem test split from \citet{lightman2023let}, consistent with \citet{snell2024scaling}.
Following their protocol, we define problem difficulty as a function of the base model's capabilities: we estimate pass@1 from 64 samples per problem and bin questions into five quintiles of increasing difficulty.
We use two-fold cross-validation on difficulty bins to avoid confounding bin selection with evaluation.
For additional validation, we include results on GSM8K~\citep{cobbe2021training} (1,319 test problems).

\paragraph{Models.}
Unless stated otherwise, the backbone generator is \textbf{Llama~3.2 1B}.
The verifier is a process reward model (PRM) trained via Monte Carlo rollouts following \citet{wang2024math}, consistent with the verifier setup in \citet{snell2024scaling}.

\paragraph{Controlled generator--verifier protocol.}
All methods share (i) the same structured move interface, (ii) the same generator $\mathrm{Gen}$ (and retrieval module, if enabled), and (iii) the same verifier $\mathcal{V}$.
Thus, differences are attributable to \emph{selection and allocation} rather than model identity.

\paragraph{Primary cost metric: verifier calls.}
We treat each verifier invocation as one unit of dominant cost and report
\begin{equation}
C_{\text{ver}} \;=\; \sum_{t=1}^{T} \big|\mathcal{Q}(w_t)\big|,
\label{eq:ver_cost}
\end{equation}
where $t$ indexes visited verification decision points (including retries/backtracking) and $\mathcal{Q}(w_t)\subseteq \mathcal{M}(w_t)$ is the subset of moves queried to the verifier at $w_t$.
For completeness, we also report \emph{total calls} (generation + verification), but conclusions are unchanged in our call-dominated regime.

\subsection{Baselines}
\label{sec:baselines}

All baselines use the same generator $\mathrm{Gen}$, the same structured move interface, and the same verifier $\mathcal{V}$.

\textbf{Best-of-$N$} samples $N$ complete solutions and selects the answer with highest aggregated verifier score, following the ``best-of-N weighted'' selection from \citet{snell2024scaling} that marginalizes across solutions with the same final answer.

\textbf{Beam Search} maintains a beam of partial states and expands each state by proposing candidate moves.
Following \citet{snell2024scaling}, we keep the top-$b$ partial states after each expansion step using PRM scores, and query $\mathcal{V}$ according to the baseline's acceptance rule.
We report the total verifier calls accumulated across the entire search. We select $b=4$.
This represents intermediate-state verification with \emph{uniform} allocation (fixed beam width throughout).

\textbf{Majority Voting} samples $N$ complete solution traces and returns the most frequent final answer among samples.
No verifier is used for selection; this tests whether verification provides value beyond sampling diversity.

\subsection{Main results: accuracy--cost frontier}
\label{sec:main_results}

Table~\ref{tab:main} reports overall accuracy and verifier-call cost on \textsc{MATH}.
Across all methods, our objective is not to maximize accuracy at any cost, but to improve the \emph{call-efficiency frontier}, i.e., achieve higher end-task correctness under fewer expensive verifier invocations.

Our approach achieves \textbf{55.2\%} accuracy while using \textbf{an average of 44.8} verifier calls.
This outperforms all solution-level sampling baselines at the same nominal budget ($64$ calls): Best-of-$N$ reaches 42.4\%, Majority Vote reaches 44.6\%, and Beam Search reaches 51.8\%.
Notably, we improve over the strongest listed baseline (Beam Search) by \textbf{+3.4} points while using \textbf{30\%} fewer verifier calls (44.8 vs.\ 64). In this setting, the gains are consistent with the hypothesis that decision difficulty is heterogeneous \emph{within} a single problem: allocating verification locally at ambiguous branching points is more effective than spending the same calls uniformly or only at the solution level.

\begin{table}[t]
\centering
\small
\begin{tabular}{lcc}
\toprule
Method & Verifier calls $\downarrow$ & Acc (\%) $\uparrow$ \\
\midrule
0-shot CoT & - & 30.6\\
Best-of-$N$ ($N$=64) & 64 & 42.4 \\
Majority Vote ($N$=64) & 64 & 44.6 \\
Beam Search ($b$=4, $N$=64) & 64 & 51.8 \\
\midrule
\textbf{Ours (gates + hybrid + state-$k$)} & \textbf{44.8} & \textbf{55.2} \\
\bottomrule
\end{tabular}
\caption{Overall performance on \textsc{MATH}. Verifier calls are the primary cost metric. Our method improves accuracy over solution-level sampling baselines while using substantially fewer verifier calls.}
\label{tab:main}
\end{table}

\subsection{Budget-matched comparison under fixed generation budgets}
\label{sec:budget_match}

To test whether intermediate-state allocation provides benefits beyond \emph{solution-level} selection, we compare several common inference-time strategies under \emph{matched verifier-call budgets}.
For a target budget $C_{\text{ver}}$, we run each method with the same number of generations per problem (so the x-axis is a shared \emph{generation budget}) and count verifier calls as the dominant cost.

\paragraph{Baselines under the same budget axis.}

\begin{figure}[t]
    \centering
    \includegraphics[width=\linewidth]{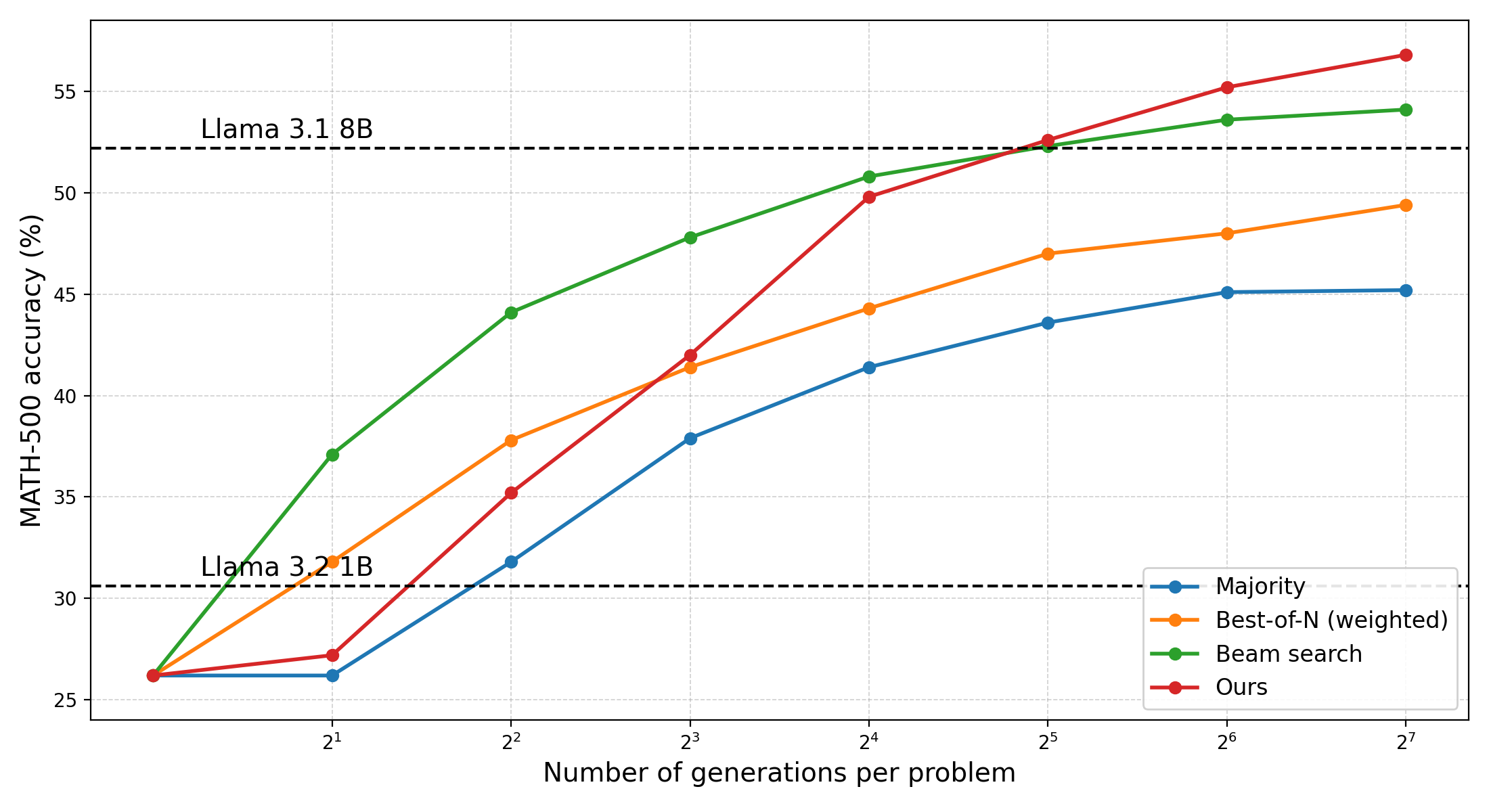}
    \caption{\textbf{Budget-matched comparison across inference strategies.} Accuracy on \textsc{MATH-500} versus number of generations per problem $N$ (x-axis). We report Majoritiy voting, solution-level Best-of-$N$ (weighted), Beam search ($b{=}4$), and our intermediate-state allocation method.}
    \label{fig:budget_match}
\end{figure}

Figure~\ref{fig:budget_match} shows accuracy as a function of generation budget $N\in\{2,4,8,16,32,64,128\}$. Across budgets, our method is consistently more call-efficient, achieving the best accuracy among budget-matched strategies, especially at moderate-to-large budgets.
Notably, beam search provides a strong improvement over solution-level selection at small-to-moderate budgets (consistent with the value of intermediate branching), but saturates earlier than our method.
This pattern supports our thesis: \emph{how} verification is distributed across intermediate states (via gating, hybrid scoring, and state-conditional $k(w)$) can materially change the accuracy achievable under a fixed verifier-call regime.

\subsection{Ablations: disentangling gating, scoring, and allocation}
\label{sec:ablation}

To isolate the effect of feasibility filtering, the ``gates-only'' condition verifies \emph{all} surviving candidates after gating, without changing the verification policy otherwise.
Subsequent ablations incrementally add scoring and adaptive allocation while keeping the same gated candidate set.

\begin{table}[t]
\centering
\small
\begin{tabular}{lcc}
\toprule
Configuration & Verifier calls $\downarrow$ & Acc (\%) $\uparrow$ \\
\midrule
Verify-all (no gates) & 64.0 & 45.0 \\
\midrule
Gates-only & 58.0 & 47.6 \\
\midrule
Gates + $D_{\text{type}}$ (fixed $k$) & 54.0 & 51.8 \\
\midrule
\textbf{Full} & \textbf{44.8} & \textbf{55.2} \\
\bottomrule
\end{tabular}
\caption{Ablations on \textsc{MATH}. ``Gates-only'' verifies all gated survivors to isolate pruning effects.}
\label{tab:ablation}
\end{table}

The result ins shown in table.~\ref{tab:ablation}. First, feasibility gating alone reduces verifier calls from 64 to 58 while \emph{improving} accuracy (45.0$\rightarrow$47.6), indicating that a substantial fraction of generated moves are deterministically invalid and waste verification budget.
Second, adding structural distance scoring further improves accuracy (47.6$\rightarrow$51.8) while reducing calls, showing that even among feasible candidates, informed prioritization matters.
Finally, state-conditional allocation provides the largest marginal gain: adaptive $k(w)$ reduces verifier calls by an additional 17\% (54.0$\rightarrow$44.8) while increasing accuracy to 55.2\%.
This confirms that uncertainty varies meaningfully across intermediate states, and that reallocating verification effort locally yields a  better accuracy--cost frontier than uniform allocation.

\subsection{Backbone and dataset variations}
\label{sec:backbone}

\begin{figure}[t]
    \centering
    \includegraphics[width=\linewidth]{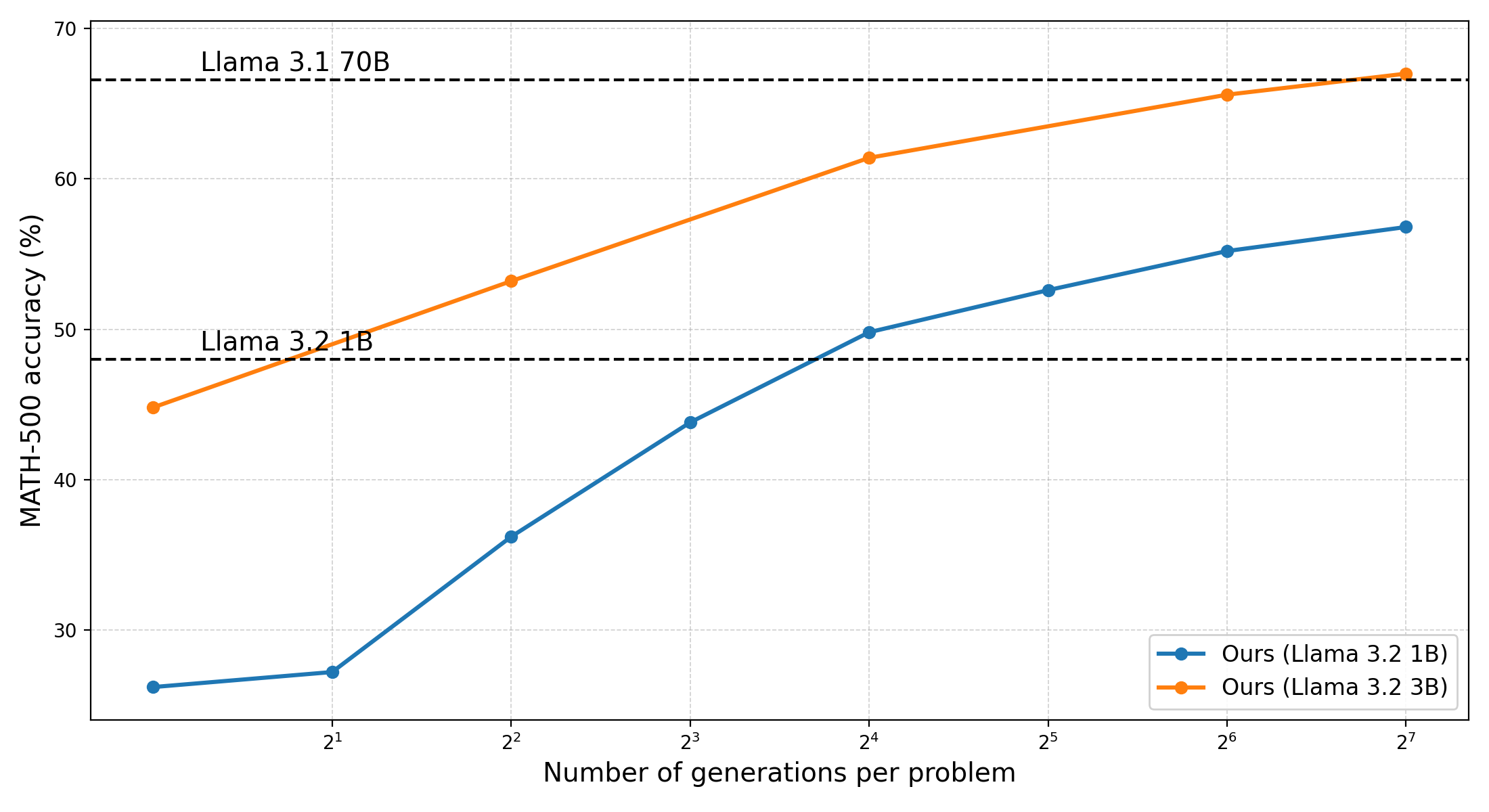}
    \caption{Backbone scaling for our framework on \textsc{MATH}-500. We plot accuracy versus the number of generations per problem (i.e., the verifier-call budget) for two backbones: Llama~3.2 1B and Llama~3.2 3B. Dashed horizontal lines indicate the 1-shot baselines of Llama~3.2 1B and Llama~3.1 70B for context.}
    \label{fig:backbone}
\end{figure}

We evaluate how backbone capacity affects the accuracy--budget curve of our \emph{state-level selective verification} policy on \textsc{MATH}-500 (Figure~\ref{fig:backbone}). Using the same framework and budget sweep, upgrading the generator from \textbf{Llama~3.2 1B} to \textbf{Llama~3.2 3B} yields a consistent improvement at every budget. In particular, the 3B backbone achieves substantially higher accuracy under small budgets (e.g., $2^2$--$2^4$ generations), and continues to improve up to $2^7$ generations, where it reaches \textbf{67.0\%} on \textsc{MATH}-500.

A key qualitative observation is that the benefit of a stronger backbone is largely \emph{complementary} to our allocation mechanism: both backbones exhibit a similar monotonic scaling trend with additional verifier budget, while the stronger backbone shifts the entire curve upward.
Moreover, the high-budget regime with Llama~3.2 3B approaches the dashed \textbf{Llama~3.1 70B} reference line, suggesting that state-level selective verification can partially substitute for model scaling by converting additional test-time budget into accuracy gains, while still benefiting from improved proposal quality as the backbone strengthens.

\section{Conclusion}

We study test-time reasoning under a verification-cost-limited setting, where the dominant marginal expense comes from calling an external verifier on intermediate steps. Our approach combines deterministic feasibility gating, hybrid pre-verification scoring, and uncertainty-aware state-level allocation, and consistently outperforms best-of-$N$, beam search, and majority voting under matched verifier-call budgets. Ablations confirm that each component contributes non-redundant gains: gating reduces wasted verification, scoring improves which candidates are worth checking, and adaptive allocation concentrates effort on ambiguous decision points. Overall, our findings suggest that finer-grained control over \emph{where} verification is applied is a key lever for improving the efficiency of test-time reasoning.

\subsection{Limitations}
\label{sec:limitations}

Selective verification cannot exceed the quality of the candidate generator.If no feasible correct move is proposed at a given decision point, no allocation or verification policy can recover the correct reasoning path. In practice, we observe this limitation most clearly on the hardest \textsc{MATH} instances (difficulty bin~5), where performance is bounded by the backbone model’s ability to propose correct intermediate steps. This mirrors the observation in \citet{snell2024scaling} that, beyond a certain difficulty regime, additional test-time compute yields diminishing returns unless proposal quality improves through larger models or better pretraining.

Feasibility gates also rely on what constraints are explicitly tracked in the context $\ell$ and on the expressiveness of the structured move interface. Constraints that are not represented in $\ell$ cannot be filtered deterministically and must be resolved by the verifier. As a result, our gates are intentionally high-precision but incomplete, and their effectiveness depends on the quality of the structured interface rather than on full semantic understanding.

\newpage

\bibliography{custom}

\appendix

\end{document}